\documentclass[journal]{IEEEtai}
    \usepackage[ruled,linesnumbered]{algorithm2e}
\usepackage{blindtext, graphicx}
\usepackage[utf8]{inputenc}
\usepackage{cite}
\usepackage{listings}
\usepackage{amssymb}
\usepackage{dcolumn}
\usepackage{siunitx}
\usepackage[breaklinks=true]{hyperref}
\usepackage{verbatim}
\usepackage{tabularx,lipsum,booktabs}
\usepackage{comment}
\usepackage{amsmath}

\usepackage{colortbl}
\usepackage[dvipsnames]{xcolor}
\usepackage[all=normal,charwidths=tight,leading=tight,
]{savetrees}
\begin{document}

\title{Complexity-Driven CNN Compression for Resource-constrained Edge AI}
\author{Muhammad~Zawish,~\IEEEmembership{Student Member,~IEEE,} Steven Davy,~\IEEEmembership{Member,~IEEE,} Lizy Abraham,~\IEEEmembership{Member,~IEEE}
        % <-this % stops a space
\thanks{M. Zawish \textbf{\textbf{(Corresponding Author)}}, S. Davy and L. Abraham are with Walton Institute for Information and Communication Systems Science, South East Technological University, Ireland, E-mails: \{muhammad.zawish, steven.davy, lizy.abraham\}@waltoninstitute.ie}

}

\maketitle

\begin{abstract}
Recent advances in Artificial Intelligence (AI) on the Internet of Things (IoT)-enabled network edge has realized edge intelligence in several applications such as smart agriculture, smart hospitals, and smart factories by enabling low-latency and computational efficiency. However, deploying state-of-the-art Convolutional Neural Networks (CNNs) such as VGG-16 and ResNets on resource-constrained edge devices is practically infeasible due to their large number of parameters and floating-point operations (FLOPs). Thus, the concept of network pruning as a type of model compression is gaining attention for accelerating CNNs on low-power devices. State-of-the-art pruning approaches, either \textit{structured} or \textit{unstructured} do not consider the different underlying nature of complexities being exhibited by convolutional layers and follow a \textit{training-pruning-retraining} pipeline, which results in additional computational overhead. In this work, we propose a novel and computationally efficient pruning pipeline by exploiting the inherent layer-level complexities of CNNs. Unlike typical methods, our proposed complexity-driven algorithm selects a particular layer for filter-pruning based on its contribution to overall network complexity. We follow a procedure that directly trains the pruned model and avoids the computationally complex ranking and fine-tuning steps. Moreover, we define three modes of pruning, namely parameter-aware (PA), FLOPs-aware (FA), and memory-aware (MA), to introduce versatile compression of CNNs. Our results show the competitive performance of our approach in terms of accuracy and acceleration. Lastly, we present a trade-off between different resources and accuracy which can be helpful for developers in making the right decisions in resource-constrained IoT environments. 
\end{abstract}
\begin{comment}

\begin{IEEEImpStatement}
The impact statement should not exceeed 150 words. This section offers an example that is expanded to have only and just 150 words to demonstrate the point. Here is an example on how to write an appropriate impact statement: Chatbots are a popular technology in online interaction. They reduce the load on human support teams and offer continuous 24-7 support to customers. However, recent usability research has demonstrated that 30\% of customers are unhappy with current chatbots due to their poor conversational capabilities and inability to emotionally engage customers. The natural language algorithms we introduce in this paper overcame these limitations. With a significant increase in user satisfaction to 92\% after adopting our algorithms, the technology is ready to support users in a wide variety of applications including government front shops, automatic tellers, and the gaming industry. It could offer an alternative way of interaction for some physically disable users.
\end{IEEEImpStatement}
\end{comment}


\IEEEpeerreviewmaketitle

\section{Introduction}
Recent years have witnessed a plethora of Internet of Things (IoT)-empowered applications due to the end users' requirements for a real-time seamless experience. Such requirements necessitate the concept of edge computing, which refers to moving Artificial Intelligence (AI) from the cloud to the network edge, which is closer to the end-users \cite{wang2021effective,zawish2022towards}. As a result, users are provided with secure and low-latency decisions since their data is now being processed on the devices which are deployed in close vicinity. IoT devices such as smartphones, surveillance cameras, and Unmanned Aerial Vehicles (UAVs) can be used to leverage edge intelligence as they are now able to deal with some amount of computations \cite{li2019edge}. However, the computational and storage resources are not always abundant on such devices; for this reason, they are known as resource-constrained devices.

Deep learning as a subbranch of AI has shown remarkable performance in various computer vision applications using extraordinary capabilities of Convolutional Neural Networks (CNNs) \cite{krizhevsky2012imagenet, simonyan2014very, he2016deep}. However, due to CNNs' nature of being highly parameterized, it becomes difficult to port them on resource-constrained edge devices where computations, storage, and power are limited. For this reason, CNNs need to be optimized in a bid to make them suitable for resource-constrained devices. Therefore, the requirement to meet the real-time application's demands running on these devices has led to an interest in model compression \cite{han2015deep}. 

Neural network pruning is a type of model compression technique that aims to effectively optimize the CNNs by eliminating redundant neurons and connections subject to the performance of a loss function. Prior works on unstructured pruning have achieved high sparsity along with the theoretically acceptable performance \cite{han2015learning, han2015deep, guo2016dynamic}. However, unstructured pruning of CNNs is not implementation friendly as it leads to irregular sparsity, which merely fulfills expected speedup and throughput in real-world applications. In contrast, many works have focused on filter pruning which shrinks the network by selectively removing the unimportant filters to reflect the structured sparsity \cite{l1norm, luo2017entropy, jordao2020deep}. As a result, filter-level pruning provides models with adequately decreased convolution filters besides realistic performance improvements. 

\begin{figure}
\includegraphics[width=0.5\textwidth]{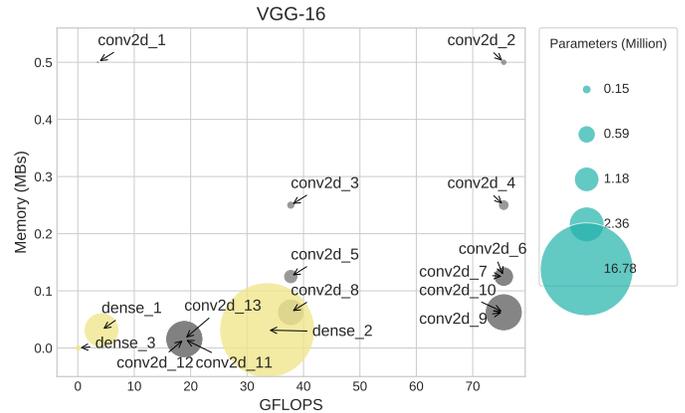}
\caption{The VGG-16 architecture consists of 13 convolutional layers and 3 fully-connected layers. Each layer has a different ratio and nature of complexity i.e. number of parameters, FLOP count, and memory size. The size of circle represents the parameter-complexity, while x-axis and y-axis shows giga FLOPs and memory based complexity respectively.}
\label{fig1}
\end{figure} 

Nevertheless, existing ranking-based pruning schemes focus only on removing unimportant filters based on some predefined importance criteria, which not only introduce computational overhead but also result in loss of performance \cite{yamamoto2018pcas,li2016pruning,jordao2020deep,wu2021pruning,zhuang2018discrimination}. To overcome this, directly training pruned target models from random initialization can achieve the same performance as the model obtained from ranking-based pruning \cite{rethinking}. Thereby, it is not necessary to start with a highly parameterized large model and then prune it; instead, one could directly train the target pruned model to achieve the same level of performance. 

Moreover, previous CNN pruning approaches do not consider the different underlying nature of complexities being exhibited by convolutional and fully connected layers. As shown in Fig \ref{fig1}., each layer of VGG-16 architecture contributes differently to each type of complexity, i.e., parameters, memory, and floating-point operations (FLOPs). Most of the parameters belong to fully connected layers, while convolutional layers possess most of the FLOPs and also occupy most of the memory size. Thus, developers must take care of the required level and nature of complexity when pruning models for specialized intelligence for resource-constrained platforms.

In this work, we address the above problems by leveraging the idea of directly training a small model, where the target model will be achieved using proposed complexity-driven pruning, and then it will be trained on the given dataset. Unlike typical ranking-based approaches, we skip the filter-ranking and fine-tuning steps due to their unnecessary computational overhead and instead adopt random pruning of filters which can also achieve comparable or even better performance\cite{random}. Our approach is subtle in a way that it gives a free hand to developers to optimize models based on the different levels of resource and budget requirements. As opposed to existing pruning methods where layers are selected for filter-pruning on an ad-hoc basis, we propose complexity-driven optimization, which considers the complexity of each layer and selects the layer automatically subject to its contribution to the overall network complexity. In the subsequent section, we present the novelty and core contributions of the proposed work and the organization of the article. 
\subsection{Novelty, Contributions and Organisation of the Proposed Work}
\begin{figure*}[h]
\includegraphics[width=1\textwidth]{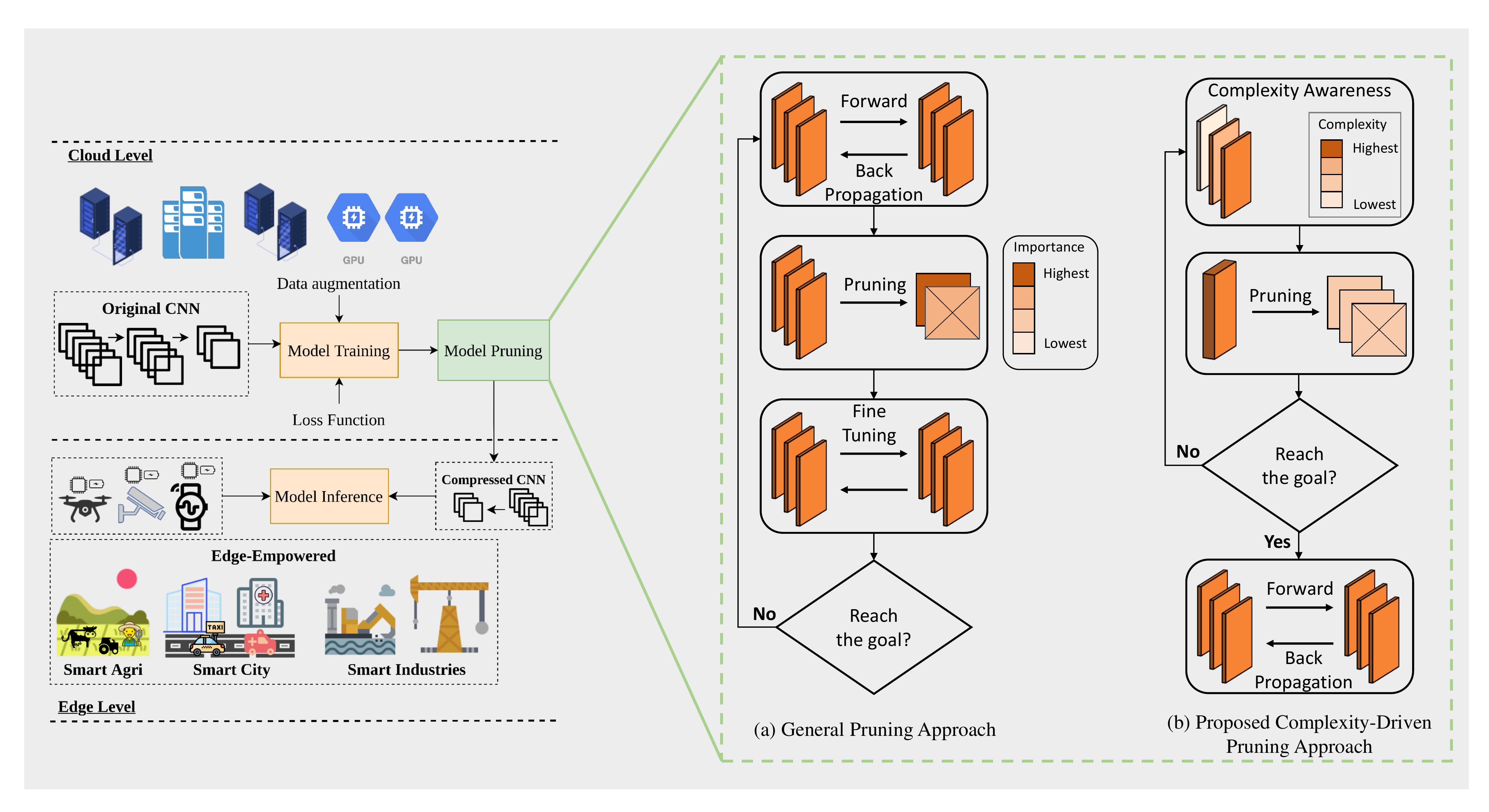}
\caption{A comparison between a typical structured pruning pipeline and the proposed complexity-driven approach. (a) shows a three-stage approach involving computationally intensive ranking and fine-tuning steps, and (b) shows a proposed complexity-driven approach skipping ranking and fine-tuning steps. }
\centering\label{fig2}
\end{figure*}

In this article, we propose a solution based on complexity-driven compression of CNNs in order to realize IoT-enabled edge-AI based applications. We summarise the key contributions of this work as follows: 
\begin{itemize}
\item We propose a novel computationally efficient framework for structured pruning where the pruned model is trained only once, avoiding the computationally-intensive ranking and fine-tuning steps. The target model is first achieved using a proposed complexity-driven pruning algorithm and then trained for a given task. The proposed scheme is illustrated in Fig. \ref{fig2} (b). 
\item While pruning filters, we consider the weight of layer-level complexity in the overall network to select a particular layer in each iteration. This justifies the overall pruning strategy as layers with more complexity will be pruned more as compared to layers exhibiting less complexity. 
\item Effectively, our approach provides versatility by providing different modes of compression such as parameter based, FLOPs-based, and memory-based. We show that using our method how accuracy can be traded off with varying types of complexities. It can be helpful for developers in repurposing the compression strategy based on the availability of resources.  
\item Moreover, we evaluate the training efficiency of our proposed approach compared to the typical \textit{training-pruning-retraining} pipeline. The results reflect the significant amount of reduction in training time on resource-rich GPUs and resource-constrained edge devices. 
\item Finally, we evaluate our proposed approach on AlexNet \cite{krizhevsky2012imagenet}, VGG-16 \cite{simonyan2014very}, ResNet-50\cite{he2016deep}, and MobileNetV2 \cite{sandler2018mobilenetv2} using CIFAR-10 and CIFAR-100 datasets \cite{krizhevsky2009learning}.  Our results are consistent and essentially provide competitive performance with lesser resource requirements than the state-of-the-art ranking-based approaches.

\end{itemize}
The rest of the paper is organized as follows. The related background on CNN compression, particularly the existing literature on pruning methods, is reviewed in Section II. Section III discusses the motivation and problem design of the proposed work, followed by experimental setup, performance evaluation, and use cases in Section IV. Finally, Section V concludes the proposed work. 

\section{Related work}
The practical requirements of CNNs have pushed the interests of researchers towards CNN compression where the objective is to either i) design lightweight and efficient networks \cite{howard2017mobilenets,ayi2020rmnv2,zhong2022dualconv}, or ii) prune weights/filters from existing CNNs \cite{han2015learning, han2015deep,guo2016dynamic, l1norm, luo2017entropy,jordao2020deep,yamamoto2018pcas,li2016pruning,wu2021pruning,zhuang2018discrimination} \cite{zhong2020merging, salama2019prune, salehinejad2021pruning, carballo2020accuracy}, or iii) replace memory-intensive weights by quantised or binary weights \cite{han2015deep,courbariaux2016binarized}, iv) or train a smaller network using a large (teacher) network as a guide \cite{hinton2015distilling}. However, subject to the scope of the paper, below we review the state-of-the-art pruning approaches.

\subsection{Weight/Neuron Pruning} Pruning weights is a type of unstructured pruning that reduces layer weights. Han et al. \cite{han2015learning} were the first to use it, removing irrelevant connections to compress the model size. In this method, the authors used the ImageNet dataset to train the model parameters, claiming compression of up to 90\% for the AlexNet model and 91\% for the VGG-16 model. Weight sharing based on unimportant connections was performed by \cite{han2015deep}, and Huffman encoding was used to decrease the weight in order to increase the compression rate. Authors in \cite{carreira2018learning} proposed a pruning approach in which unimportant weights are located, and it was demonstrated that loss changes decrease when weights are compressed. MSN\cite{zhong2020merging} prunes the models in a single-shot manner by first clustering the similar neurons and then merging them to produce sparse structures. While authors in \cite{lin2019towards} propose an iterative approach using generative adversarial learning to learn the sparse soft mask, which forces the output of specific structures to be $zero$. However, all these approaches provide a model with high sparsity, which results in the increased complexity of hyperparameter optimization. 

\subsection{Filter Pruning} Pruning filters is a type of structured pruning that was introduced to overcome the loopholes of sparsity-induced approaches. The core idea is to achieve structured sparsity in CNNs by eliminating the parts of the structure, such as filters. For instance, \cite{l1norm} proposed the use of \textit{l}\textsubscript{1}-norm to rank the importance of filters, so that the filters with lower \textit{l}\textsubscript{1}-norm can be removed as their contribution is insignificant. Alternatively, authors in \cite{luo2017entropy} replaced \textit{l}\textsubscript{1}-norm with entropy as a measure to calculate filter importance. In this case, the higher entropy of a certain filter reflects its significance in terms of increasing accuracy. In \cite{jordao2020deep}, authors proposed the use of partial least squares to capture the relationship of filter importance with class importance in a low dimensional space. As opposed to single-shot pruning approaches, \cite{jordao2020deep} introduced an iterative manner of pruning where the model is pruned iteratively with a small pruning ratio instead of pruning once with a large pruning ratio. Similarly, authors in \cite{salehinejad2021pruning} propose a rising energy model to quantify the inactivity of convolutional kernels in order to prune them iteratively. In contrast to\cite{salehinejad2021pruning,jordao2020deep}, authors in \cite{carballo2020accuracy} propose a relatively simple approach to assign pruning priority to each layer according to its impact on accuracy.

\subsection{Pruning from Scratch}
The approaches discussed above follow a three-stage pipeline: training a CNN, pruning based on some ranking criteria, and fine-tuning to regain accuracy. However, authors in \cite{random} claimed to achieve similar performance by replacing the filter-ranking part with random pruning of filters. It is observed that CNNs have a characteristic of plasticity - they can recover the damage in the fine-tuning step no matter which filters are pruned. Similarly, \cite{rethinking} claimed that retraining/fine-tuning the pruned models on a large number of epochs requires additional computing costs. Thus, they proposed the idea of training only once, where directly training a small model can achieve competitive performance as other approaches \cite{rethinking}. 

None of the aforementioned studies considers the underlying complexity proportion of a CNN layer, and they either prune the model layer by layer or all layers at once in a single iteration. This does not justify the pruning strategy, as each layer possesses a different level and nature of complexity. To overcome the above problems, our work takes inspiration from \cite{random} and \cite{rethinking}, where we directly train a small pruned model on a given dataset. The small model is produced using complexity-driven pruning, which randomly prunes filters from a certain layer based on its relative weight. The proposed work is able to achieve competitive performance as state-of-the-art techniques on AlexNet \cite{krizhevsky2012imagenet}, VGG-16 \cite{simonyan2014very}, ResNet-50\cite{he2016deep}, and MobileNetV2 \cite{sandler2018mobilenetv2} using CIFAR-10 and CIFAR-100 datasets \cite{krizhevsky2009learning}.\\ 

\section{Proposed Approach}
In this section, we discuss the motivation of our approach, followed by the details on proposed complexity-driven pruning, and a tutorial detail of algorithm. 
\subsection{Motivation}
A typical CNN consists of an arbitrary number of convolutional and fully connected layers stacked up on top of each other which account for the overall CNN complexity and performance. A typical convolutional layer $k = Conv(X, W)$ receives an input tensor $X\in\mathbb{R}^{C_{in} \times W \times H }$ and applies a trainable tensor of weights $W \in \mathbb{R}^{C_{out} \times C_{in} \times K_w \times K_h} $ to produce an output tensor $O \in \mathbb{R}^{C_{out} \times W \times H } $,
\begin{equation}
    Conv(X, W)_i= \sum_{c\in[C_{in}]} W_{i,c} * X_c \forall i \in [C_{out}],
\end{equation}

where $C_{in}$ and $C_{out}$ denote the number of input and output channels or filters, respectively. Such convolution operations are responsible for most of the computations inside CNNs during training and inference. Therefore, to enable them for resource-constrained execution, prior works on structured pruning rely on a three-stage pipeline, as shown in Fig 2 (a). At first, a highly parameterized network is trained, then filters are ranked and pruned from one layer or across all layers based on some filter-importance criteria, and lastly, the network is fine-tuned on a specific task to minimize the loss. In contrast, as shown in Fig 2 (b), we advance the spirit of directly training the compressed model, where the target model is derived using complexity-driven pruning and then trained only once for a given task. This idea can achieve comparable or even better performance by avoiding the complex ranking and fine-tuning steps that account for a high computational cost in the whole iterative pruning process. Firstly, given a predefined CNN topology, our approach achieves the pruned small model by exploiting the intrinsic layer-level complexities, which are in the form of layer parameters, layer-FLOPs, and layer-memory. A particular layer is selected in each pruning iteration for random filter pruning of the predefined percentage of filters. Once the pruned target architecture is achieved, then it is trained on the given dataset. Moreover, our technique helps developers to prune the model based on the available resources. For instance, if the objective is to minimize only the computations while sacrificing the storage, then the pruning mode can be selected as FLOPs and vice versa. 
\begin{figure*}
\includegraphics[width=1\textwidth]{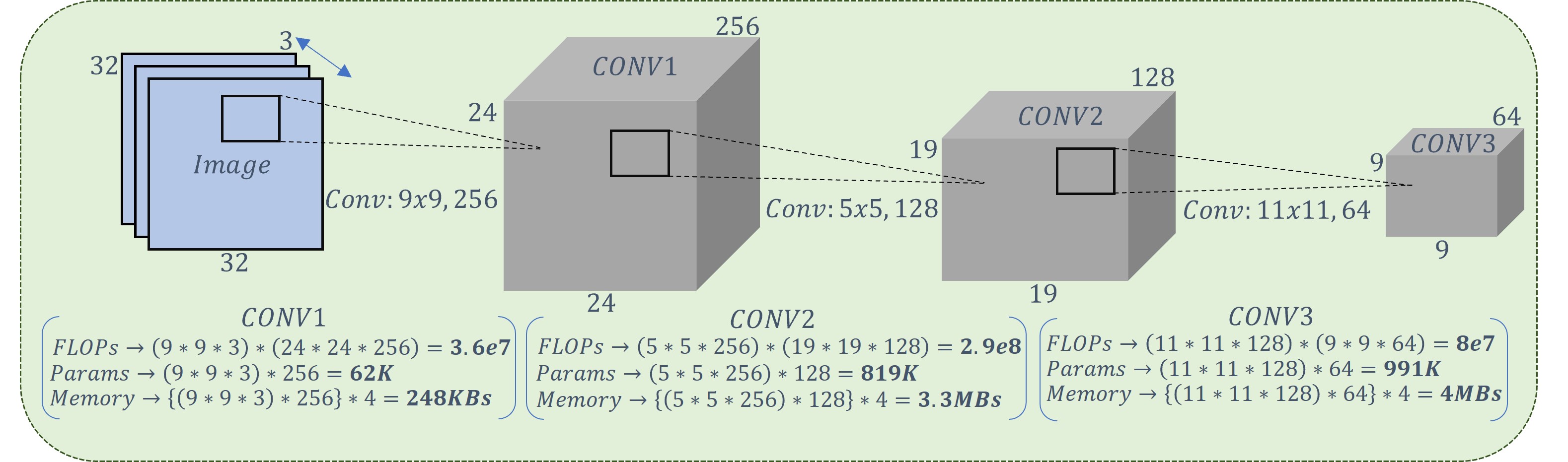}
\caption{An example of a CNN with 3 conv layers with uneven distribution of filters showing the calculation of FLOPs, parameters, and memory. Each layer exhibits the different number of parameters, FLOPs, and memory requirements as we move from upper layers to lower layers. The spatial dimensions are calculate using: $(Input \; size - Filter\; size)/stride+1$. Note that we have considered $stride=0$ at this stage.}
\label{fig3}
\end{figure*} 
\subsection{Complexity-Driven Pruning}
Suppose a CNN’s topology consists of $\mathit {K}$ convolutional layers, and the $\mathit {k}$-th layer has $\mathit {N_k}$ filters. For all filters $\mathit {N_k}$ of $\mathit {k}$-th layer, we encode the parameters of each filter as  $W_{k}^{n}\in\mathbb{R}^{C_{k}^{in} \times \Omega_k \times \Omega_k }$, where $C_{k}^{in}$ denotes number of input channels, and $\Omega$ denotes size of the kernel. Thereby, combining all filters together, we get the filter parameters set $W = \{\{W_{n}^{k}\}_{n=1}^{N_k}\}_{k=1}^{K}$, where $N = \sum\limits_{k\in K} N_k$ denotes the total number of filters in the entire CNN. Then, for a dataset $D = \{x_i, y_i\}_{i=1}^Z$ with $Z$ inputs and their corresponding labels, the pruning mechanism aims to find a CNN model $M^{'}$ with fewer parameters as compared to the baseline CNN model $M$ using Eq. \ref{eq2}

\begin{equation} \label{eq2}
    \min_{\hat{W}} \ \ L(D|\hat{W}) \ \ \text{s.t.} \ \ || \hat{W} ||_{0}  \leq || W ||_{0}
\end{equation}

where $L(D|\hat{W})$ denotes the loss between model predictions and ground truth, and $\hat{W}$ denotes the number of remaining parameters in the model $M^{'}$. 

\begin{algorithm}
\caption{Complexity-Driven Pruning}
\label{algo1}
\SetAlgoLined

Input CNN topology $\mathit{M}$ with $\mathit{K}$ convolutional layers 

Input required complexity $\gets \{\mathit C_r\}$ 

Input mode $\gets \{\mathit FLOPs, MEMORY, PARAMs\}$

Input Pruning ratio $P_r$

\Begin{
    \Switch{mode}{
        \Case{FLOPs}{
            \While{$C_{M} \leq C_r$}{
                $\mathit{weights = [ ]}$
                
                \For{$k\in\{1,...,K\}$}{
                    $\omega_{k}^{f}= C_{in}^{k}\times (\Omega^k)^2 \times  C_{out}^{k}\times S_{out}^{k}$
                    
                    $\mathit{weights\mathrel{{+}{=}}\frac{\ \omega_k^f}{\sum\limits_{k\in K} \omega_k^f}}$
                }
                $\mathit {k} = \mathit sample(K, weights)$
  
                $\mathit {k} = \mathit prune\_filters(k, P_r)$
            }
        }
        
        \Case{MEMORY}{
            \While{$C_{M} \leq C_r$}{
                $\mathit{weights = [ ]}$
                
                \For{$k\in\{1,...,K\}$}{
                     $\omega_{k}^{m}= C_{in}^{k}\times (\Omega^k)^2 \times  C_{out}^{k} \times 4$
                    
                    $\mathit{weights\mathrel{{+}{=}}\frac{\ \omega_k^m}{\sum\limits_{k\in K} \omega_k^m}}$
                }
                $\mathit {k} = \mathit sample(K, weights)$
  
                $\mathit {k} = \mathit prune\_filters(k, P_r)$
            }
        }
        \Case{PARAMs}{
            \While{$C_{M} \leq C_r$}{
                $\mathit{weights = [ ]}$
                
                \For{$k\in\{1,...,K\}$}{
                    $w_{k}^{p}= C_{in}^{k}\times (\Omega^k)^2 \times  C_{out}^{k}$
                    
                    $\mathit{weights\mathrel{{+}{=}}\frac{\ \omega_k^p}{\sum\limits_{k\in K} \omega_k^p}}$
                }
                $\mathit {k} = \mathit sample(K, weights)$
                $\mathit {k} = \mathit prune\_filters(k, P_r)$
            }
        }
           
    }
    Output the pruned CNN model $\mathit {M^{'}}$ for training
}
\end{algorithm}

In each iteration, typical structured pruning focuses on identifying the least important filters and prunes a certain percentage of filters from either all layers or a single layer. However, this approach is not justified as filters are unevenly distributed across a network's lower and upper layers, hence resulting in a different number of parameters, FLOPs, and memory requirements. In Fig. \ref{fig3}, we show an example in support of this argument representing the calculation of complexities in a 3-layered CNN architecture. It can be seen that inconsistency in the number of filters highly correlates with inconsistency in type and scale of complexity. Thus, it is unfair to prune the same ratio of filters from both a layer with lower complexity and a layer with higher complexity, as it will degrade the performance of the network. To overcome this, we propose the idea of a complexity-driven selection of layers in each pruning iteration using the weighted random sampling technique. Thereby, the likelihood of selection of layer $\mathit {k}$ in each round is according to the complexity weight associated with it. Let's denote the complexity weight of layer $\mathit {k}$ be $\mathit {\omega_k}$ then the probability $\mathit {P}$ of $\mathit {k}$ to be randomly sampled is proportional to its relative weight i.e. $\mathit {P_k}=\frac{\omega_k}{\sum\limits_{k\in K} \omega_k}$. Note that the complexity weight $\mathit {\omega_k}$ can be the number of FLOPs, parameters, or memory size. We show the overall pruning scheme in Algorithm \ref{algo1}. To make the right trade-off between performance and resource requirements, we explore different types of complexities possessed by each layer. 

\subsubsection{FLOPs-Aware Pruning}
Floating-point operations are the computational bottleneck in a CNN which are influenced by convolutional layers as they involve matrix multiplications. As observed in Fig. \ref{fig1}, the lower layers of VGG-16 consume a large number of total FLOPs due to a high-resolution input image. In terms of workload, the complexity of the model affects how much it takes to run because larger models require more processors and will be more (computationally) expensive. Moreover, denser models lead to a higher duty cycle, which in turn results in more extended periods that the processor will spend running rather than idle. This results in increased power consumption and heat exhaustion. Therefore, to accelerate the inference of a CNN on a computationally constrained device, it is essential to prune lower layers more as compared to upper layers. For each convolutional layer $\mathit {k}$, its FLOPs-based complexity can be calculated using Eq. \ref{eq3}, and FLOPs-driven pruning is formulated using Eq. \ref{eq4}

\begin{equation} \label{eq3}
   	\omega_{k}^{f}= C_{in}^{k}\times (\Omega^k)^2 \times  C_{out}^{k}\times S_{out}^{k},
\end{equation}

\begin{equation}\label{eq4}
\min \sum_{k=1}^{K} \omega_k^f \ \ \text{s.t.} \ \ \forall k \ \mathit {P_k}=\frac{\ \omega_k^f}{\sum\limits_{k\in K} \omega_k^f}
\end{equation}

\subsubsection{Memory-Aware Pruning}

The reduction in memory of a CNN is critical when the aim is to achieve both computation and energy efficiency. Since a small model size consists of not only fewer FLOPs but also lesser Dynamic Random-Access Memory (DRAM) traffic involving the read and write of feature maps and model parameters. This mode of pruning is often required for compressing CNNs for microcontrollers, as they have a restricted storage budget. Moreover, in order for the deep models to run at the edge, they must fit within the target device's RAM without disrupting the IoT application at the runtime. To achieve this, the memory-based complexity of each convolutional layer $\mathit {k}$ can be calculated using Eq. \ref{eq5}, and memory-driven pruning is formulated using Eq. \ref{eq6}

\begin{equation}\label{eq5}
    \omega_{k}^{m}= C_{in}^{k}\times (\Omega^k)^2 \times  C_{out}^{k} \times 4
\end{equation}

\begin{equation}\label{eq6}
\min \sum_{k=1}^{K} \omega_k^m \ \ \text{s.t.} \ \ \forall k \ \mathit {P_k}=\frac{\ \omega_k^m}{\sum\limits_{k\in K} \omega_k^m}
\end{equation}

\definecolor{Gray}{gray}{1}
\definecolor{Gray1}{gray}{0.94}
\definecolor{Gray2}{gray}{0.88}

\begin{table*}[ht]
\caption{Comparison with state-of-the-art approaches on VGG-16 and CIFAR-10. Accuracy shows the gain in accuracy, where negative sign denotes the loss regarding original network.  $\downarrow$ denotes the reduction in percentage w.r.t the unpruned network.}
\label{tab1}
%%\centering % redundant
%%\renewcommand\footnoterule{\kern -1ex} % What is this instruction doing here??
\renewcommand{\arraystretch}{1.3}
\setlength\tabcolsep{0pt} % make LaTeX figure out the intercolumn separation
\begin{tabular*}
{\linewidth}{@{\extracolsep{\fill}} l *{8}{c}} % '1.4', not '6.5'

\toprule 
\rowcolor{Gray2}\textbf{Approach} & \multicolumn{4}{c}{\textbf{Baselines}} & \multicolumn{4}{c}{\textbf{Post-Pruning}}\\
\cmidrule{2-5} \cmidrule{6-9} &
Accuracy (\%) & FLOPs & 
Memory (MBs) & Parameters (Millions) & 
Accuracy (\%) & FLOPs (\%)$\downarrow$ & 
Memory (\%)$\downarrow$ & Parameters (\%)$\downarrow$ \\
\midrule 
 MSN \cite{zhong2020merging} & $93.62$ & -- & -- & -- & $0.04$ & $38.05$ & -- & $86.09$ \\ 
GAL \cite{lin2019towards} & $93.94$ & -- & -- & -- & $-3.36$ & $45.2$ & -- & $82.2$ \\ 
PLS \cite{jordao2020deep} (Itr=1) & 87.05 & $6.65\times10^8$ & $127$ & $33.64$ & $0.46$ & $28$ & $18.38$ & $16$ \\ 
PLS \cite{jordao2020deep} (Itr=5) & 87.05 & $6.65\times10^8$ & $127$ & $33.64$ & $-1.08$ & $67$ & $44.51$ & $30$ \\ 
PLS \cite{jordao2020deep} (Itr=10) & 87.05 & $6.65\times10^8$ &  $127$ & $33.64$ & $-9.7$ & $88$ & $62.9$ & $37.62$ \\

\hline
\textbf{Our (PA)} & $90.96$ & $6.61\times10^8$ &  $125$ & $33.19$ & $\textbf{1.76}$ & $45$ & $49.09$ & $\textbf{84.25}$ \\ 
\textbf{Our (FA)} & $90.96$ &$6.61\times10^8$ &  $125$ & $33.19$ & $\textbf{-3.45}$ & $\textbf{89.67}$ & $62.19$ & $41.20$ \\ 
\textbf{Our (MA)} & $90.96$ & $6.61\times10^8$ &  $125$ & $33.19$ & $\textbf{-8.76}$ & $81.13$ & $\textbf{91.02}$ & $49.15$ \\ 
\bottomrule
\end{tabular*}
\end{table*}

\begin{table*}[ht]
\caption{Comparison with state-of-the-art approaches on MobileNetV2 and CIFAR-10. Accuracy shows the gain in accuracy, where the negative sign denotes the loss regarding the original network. $\downarrow$ denotes the reduction in percentage w.r.t the unpruned network.}
\label{tab2}
%%\centering % redundant
%%\renewcommand\footnoterule{\kern -1ex} % What is this instruction doing here??
\renewcommand{\arraystretch}{1.3}
\setlength\tabcolsep{0pt} % make LaTeX figure out the intercolumn separation
\begin{tabular*}{\linewidth}{@{\extracolsep{\fill}} l *{8}{c}} % '1.4', not '6.5'

\toprule 
\rowcolor{Gray2}\textbf{Approach} & \multicolumn{4}{c}{\textbf{Baselines}} & \multicolumn{4}{c}{\textbf{Post-Pruning}}\\
\cmidrule{2-5} \cmidrule{6-9} &
Accuracy (\%) & FLOPs & 
Memory (MBs) & Parameters (Millions) & 
Accuracy (\%) & FLOPs (\%)$\downarrow$ & 
Memory (\%)$\downarrow$ & Parameters (\%)$\downarrow$ \\
\midrule 
FEAM \cite{wu2021pruning}  & $74.3$ & $1.7\times10^9$ & --& $5.73$ & $-12.6$ & $25.6$ & -- & $21.11$ \\ 
RMNv2 \cite{ayi2020rmnv2} & $94.3$ & -- &  $9.14$ & $2.2$ & $-2.01$ & -- & $52.7$ & $52.2$ \\
 WM \cite{zhuang2018discrimination} & $94.47$ & -- & -- & -- & $-0.30$ & $26$ & -- & -- \\ 
DCP \cite{zhuang2018discrimination} & $94.47$ & -- & -- & -- & $0.22$ & $26$ & -- & -- \\ 
NPPM \cite{gao2021network}  & $94.23$ & -- & -- & -- & $0.52$ & $47$ & -- & -- \\

\hline
\textbf{Our (PA)} & $89.87$ & $4.92\times10^6$ &  $17.7$ & $1.36$ & $\textbf{0.07}$ & $33.05$ & $49.12$ & $\textbf{70.03}$ \\
\textbf{Our (FA)} & $89.87$ &$4.92\times10^6$ &  $17.7$ & $1.36$ & $\textbf{0.23}$ & $\textbf{50.77}$ & $28.62$ & $18.5$ \\
\textbf{Our (MA)} & $89.87$ & $4.92\times10^6$ &  $17.7$ & $1.36$ & $\textbf{0.96}$ & $39.48$ & $\textbf{53.5}$ & $30.72$ \\ 
\bottomrule
\end{tabular*}
\end{table*}

\subsubsection{Parameter-Aware Pruning}
In a typical CNN, the filters in convolutional layers account for fewer parameters than the fully connected layers. For instance, in VGG-16, 90\% of parameters belong to three fully connected layers, while 10\% belong to 13 convolutional layers. Thus, if it is required for a particular application to reduce the parameters, then the proposed technique can be used in the PARAMS mode. In this mode, the parameter space of convolutional layers can be reduced to lower not only the memory footprint but also the operational cost during the inference. To achieve this,  for each convolutional layer $\mathit {k}$, its parameter-based complexity can be calculated using Eq. \ref{eq7}, and parameter-driven pruning is formulated using Eq. \ref{eq8}

\begin{equation} \label{eq7}
     w_{k}^{p}= C_{in}^{k}\times (\Omega^k)^2 \times  C_{out}^{k}
	\end{equation}
	
	\begin{equation}\label{eq8}
 \min \sum_{k=1}^{K} \omega_k^p \ \ \text{s.t.} \ \ \forall k \ \mathit {P_k}=\frac{\ \omega_k^p}{\sum\limits_{k\in K} \omega_k^p}
	\end{equation}

where $\mathit {K}$ is the number of convolutional layers, $\mathit {k}$ denotes the layer index, $C_{in}$ is the number of input channels, $C_{out}$ is the number of output channels,  $\Omega^2$ is the filter size, $S_{out}$ is the feature size of the output layer, and $P_k$ denotes the likelihood of layer $k$ to be selected for pruning. For memory-aware, the 4 refers to the number of bytes required to store in a 32-bit system.

\subsection{Algorithm Description}
This subsection entails the description of pseudo-code presented in Algorithm \ref{algo1}. $Lines \; 1-4$ defines the input parameters required for the technique, $M$ is the unpruned model, $C_r$ is the required complexity (such as the number of parameters, FLOPs, or memory), mode ensures the objective of layer-selection, and $P_r$ is the \% of filters need to be pruned from a selected $conv$ layer. We follow a $switch-case$ design scenario in the algorithm to define the pruning strategy for each mode/case, i.e., FLOPs, MEMORY, or PARAMs. For instance, let's consider the case of FLOPs, $line \;8$ starts a $while$ loop which will includes initialisation of a list of $\mathit {weights = [ ]}$  to store the relative complexity weight of each layer $k\in\{1,...,K\}$ on $line\;9$ followed by a $for$ loop on $line\;10$. Inside $for$ loop, we iterate through each $conv$ layer, and compute its complexity $\omega_{k}^{f}$ using Eqn. 3, and append its relative weight $  \frac{\omega_{k}^{f}}{\sum\limits_{k\in K} \omega_{k}^{f} } $ into $\mathit {weights}$. Then, $line\;14$ sample layer $\mathit {k}$ for pruning using weighted random sampling, and $line\; 15$ prunes layer $k$ according to $P_r$. The $while$ loop continues $until$ it meets the following condition: $C_{M} \leq C_r$, meaning that the complexity of CNN model $M$ must be either less than or equal to the desired complexity $C_r$. Similarly, $lines\;18-28$ applies for memory-aware pruning (using Eqn. 5), and $lines\;29-39$ applies for parameter-aware pruning (Eqn. 7). Lastly, $line\;41$ produces the pruned output model ${M^{'}}$ for training.

\section{Experiments and Results}
This section consists of 4 subsections: i) evaluation of VGG-16 and MobileNetV2 on the CIFAR-10 dataset, ii) evaluation of AlexNet and ResNet-50 on the CIFAR-100 dataset, iii) resource-accuracy trade-off, and iv) the training efficiency of the proposed approach. In subsections I and II, we show a comparison of our approach with state-of-the-art methods in terms of accuracy, FLOPs, memory, and parameters of compressed models. To evaluate the suitability of the proposed approach, subsection IV highlights the impact of the proposed approach on energy consumption, latency, CPU, and memory utilization. 

We implemented the networks using Keras deep learning API with Tensorflow as a backend on the Nvidia Tesla K20 GPU with 8GB memory. The OS was Ubuntu 18.04, with Python version 3.6.12, Keras version 2.2.4, and Tensorflow version 1.15. We trained the networks for 150 epochs with a mini-batch size of 128 and a learning rate of 10\textsuperscript{-3} using an RMSprop optimizer. We performed all the experiments on 5 different seeds and averaged the results over them. For resource-constrained benchmarking, we simulated the models on an OpenStack virtual machine (VM) with 2 CPUs, 4GB of RAM, and 10GB of the hard disk. To avoid conflicts, we used the same OS and versions of Python, Keras, and Tensorflow. We benchmarked the results on 100 random images and showed the average latency and resource utilization.

\begin{table*}[ht]
\caption{Comparison with state-of-the-art approaches on AlexNet and CIFAR-100. Accuracy shows the gain in accuracy, where the negative sign denotes the loss regarding the original network.  $\downarrow$ denotes the reduction in percentage w.r.t the unpruned network.}
\label{tab3}
%%\centering % redundant
%%\renewcommand\footnoterule{\kern -1ex} % What is this instruction doing here??
\renewcommand{\arraystretch}{1.3}
\setlength\tabcolsep{0pt} % make LaTeX figure out the intercolumn separation
\begin{tabular*}{\linewidth}{@{\extracolsep{\fill}} l *{8}{c}} % '1.4', not '6.5'

\toprule 
\rowcolor{Gray2}\textbf{Approach} & \multicolumn{4}{c}{\textbf{Baselines}} & \multicolumn{4}{c}{\textbf{Post-Pruning}}\\
\cmidrule{2-5} \cmidrule{6-9} &
Accuracy (\%) & FLOPs & 
Memory (MBs) & Parameters (Millions) & 
Accuracy (\%) & FLOPs (\%)$\downarrow$ & 
Memory (\%)$\downarrow$ & Parameters (\%)$\downarrow$ \\
\midrule 
 IEM \cite{salehinejad2021pruning} & $83.31$ & -- & -- & $57.4$ & $-4.34$ & -- & -- & $35.08$ \\ 
SFP \cite{carballo2020accuracy} & $72.86$ & $710 \times 10^6$ & -- & -- & $-0.36$ & $40.98$ & -- & -- \\ 
PLS \cite{jordao2020deep} (Itr=1) & $83.06$ & $2.29 \times10^8$ & $129$ & $33.95$ & $0.17$ & $13.46$ & $4.11$ & $2.15$ \\ 
PLS \cite{jordao2020deep} (Itr=5) & $83.06$ & $2.29 \times10^8$ & $129$  & $33.95$ & $-2.7$ & $45.38$ & $17.92$ & $8.44$ \\
PLS\cite{jordao2020deep} (Itr=10) & $83.06$ & $2.29 \times10^8$ & $129$  & $33.95$ & $-4.1$ & $62.79$ & $40.61$ & $18.97$ \\

\hline
\textbf{Our (PA)} & $90.28$ & $2.29 \times10^8$ & $129$ & $33.95$ & $\textbf{0.43}$ & $20.11$ & $17.66$ & $\textbf{63.92}$ \\ 
\textbf{Our (FA)} & $90.28$ & $2.29 \times10^8$ & $129$ & $33.95$ & $\textbf{-1.15}$ & $\textbf{79.28}$ & $61.59$ & $47.06$ \\
\textbf{Our (MA)} & $90.28$ & $2.29 \times10^8$ & $129$ & $33.95$ & $\textbf{-3.06}$ & $71.90$ & $\textbf{83.31}$ & $51.26$ \\ 
\bottomrule
\end{tabular*}
\end{table*}

\begin{table*}[ht]
\caption{Comparison with state-of-the-art approaches on ResNet-50 and CIFAR-100. Accuracy shows the gain in accuracy, where the negative sign denotes the loss regarding the original network.  $\downarrow$ denotes the reduction in percentage w.r.t the unpruned network.}
\label{tab4}
%%\centering % redundant
%%\renewcommand\footnoterule{\kern -1ex} % What is this instruction doing here??
\renewcommand{\arraystretch}{1.3}
\setlength\tabcolsep{0pt} % make LaTeX figure out the intercolumn separation
\begin{tabular*}{\linewidth}{@{\extracolsep{\fill}} l *{8}{c}} % '1.4', not '6.5'

\toprule 
\rowcolor{Gray2}\textbf{Approach} & \multicolumn{4}{c}{\textbf{Baselines}} & \multicolumn{4}{c}{\textbf{Post-Pruning}}\\
\cmidrule{2-5} \cmidrule{6-9} &
Accuracy (\%) & FLOPs & 
Memory (MBs) & Parameters (Millions) & 
Accuracy (\%) & FLOPs (\%)$\downarrow$ & 
Memory (\%)$\downarrow$ & Parameters (\%)$\downarrow$ \\
\midrule 
 DualConv \cite{zhong2022dualconv} & $78.55$ & $1.3 \times 10^9$ & -- & $34$ & $-0.7$ & $29$ & -- & $26.5$ \\ 
SANet \cite{hacene2021attention} & $78$ & $1.3 \times 10^9$ & -- & $16.9$ & $-0.51$ & $80$ & -- & $76$ \\ 
PCAS \cite{yamamoto2018pcas}  & $74.46$ & $1.4 \times 10^9$ & -- & $17.1$ & $-0.83$ & $66.47$ & -- & $76$ \\ 
Pruned-B \cite{li2016pruning}  & $74.46$ & $1.4 \times 10^9$ & -- & $17.1$ & $-1.15$ & $56.28$ & -- & $54.2$ \\

\hline
\textbf{Our (PA)} & $79.31$ & $1.51\times 10^8$ & $90$ & $23.6$ & $\textbf{-0.25}$ & $32.21$ & $41.59$ & $\textbf{53.4}$ \\ 
\textbf{Our (FA)} & $79.31$ & $1.51\times 10^8$ & $90$ & $23.6$ & $\textbf{0.34}$ & $\textbf{51.33}$ & $42.7$ & $35.15$ \\ 
\textbf{Our (MA)} & $79.31$ & $1.51\times 10^8$ & $90$ & $23.6$ & $\textbf{2.23}$ & $19.38$ & $\textbf{30.2}$ & $18.1$ \\ 
\bottomrule
\end{tabular*}
\end{table*}

\subsection{Evaluation on CIFAR-10}
In this subsection, we evaluate the VGG-16 and MobileNetV2 on the CIFAR-10 Dataset. There are 60,000 tiny images in the CIFAR-10 dataset categorized into 10 distinct classes with the dimensions of 32x32. Each class has 6000 images, of which 5000 belong to the training set, while 1000 belong to the test set.  
\subsubsection{VGG-16}
Originally, VGG-16 was proposed for the ImageNet dataset, but several works have also reported its substantial performance on CIFAR-10. The architecture of the VGG-16 model is composed of 12 convolutional layers followed by 3 fully connected layers stacked on top of each other. The last layer represents the number of classes in a particular dataset, e.g., 10 for CIFAR-10, and it possesses softmax as an activation function, while ReLu was used for the rest of the layers. On CIFAR-10, we compare our approach with MSN\cite{zhong2020merging}, GAL\cite{lin2019towards}, and PLS\cite{jordao2020deep}. MSN prunes the models in a single-shot manner by first clustering the similar neurons and then merging them to produce sparse structures. While GAL and PLS are iterative approaches, GAL \cite{lin2019towards} applied generative adversarial learning to learn the sparse soft mask, which forces the output of specific structures to 0, and PLS \cite{jordao2020deep} applies partial least squares to identify the importance of filters with respect to the class labels.

We pruned VGG-16 architecture in three different modes, i.e., parameter-aware (PA), flops-aware (FA), and memory-aware (MA), using the proposed approach described in section 3. It can be observed from Table \ref{tab1} that maximum compression, i.e., $91.02\%$, is achieved with the MA mode, and maximum gain in accuracy, i.e., $1.76\%$, was observed with PA mode. This compression level is significant regarding CNN acceleration on the microcontrollers where storage is sparse. Since some applications can not bear the loss of accuracy, then PA mode can be adopted, as it reduced 84.25\% parameters while gaining 1.76\% of accuracy. In contrast, the single-shot learning-based approach MSN \cite{zhong2020merging} gained only $0.04\%$ accuracy by lowering the approximately same number of parameters as ours. The iterative process by \cite{jordao2020deep} and GAL\cite{lin2019towards} dropped $9.7\%$ and $3.36\%$ of accuracy, respectively, by reducing the approximately same number of parameters, FLOPs, and memory.  
%Similarly, in FA mode, our approach minimised the operations by 89.67\% while losing 3.45\% accuracy as opposed to the iterative approach which lost 9.7\% accuracy while minimising lesser operations than ours. 
\subsubsection{MobileNetV2}
In MobileNets\cite{howard2017mobilenets, sandler2018mobilenetv2}, the standard convolution layer is replaced with a depthwise separable convolutional layer to make them compact as compared to AlexNet or VGG-16. MobileNetV2\cite{sandler2018mobilenetv2} is more light-weight than the MobileNetV1\cite{howard2017mobilenets}, hence it contains less number of redundant parameters. However, it is still a bottleneck for small datasets such as CIFAR-10, as redundancy remains approximately the same. Thus, it is imperative to prune it for acceleration on resource-constrained devices. The architecture of MobileNetV2 is composed of 17 residual blocks with skip connections, followed by a $1\times1$ $conv$ layer, a global average pooling layer, and a softmax layer. 

We compare the performance of our approach on MobileNetV2 with  FEAM\cite{wu2021pruning}, RMNv2\cite{ayi2020rmnv2}, WM\cite{zhuang2018discrimination}, DCP\cite{zhuang2018discrimination}, and NPPM\cite{gao2021network}. FEAM prunes the filters iteratively by ranking their importance in terms of feature extraction ability. Similarly, \cite{zhuang2018discrimination} and \cite{gao2021network} are also based on structural pruning. NPPM \cite{gao2021network} uses a performance prediction network in the pruning process as a proxy of accuracy, while \cite{zhuang2018discrimination} takes both classification loss and norms into account as part of the pruning process. In contrast, RMNv2 performs architectural modifications by inducing the heterogeneous kernel-based convolutions and mish activations to make the MobileNetV2 even lighter. We show the comparative analysis in Table \ref{tab2}. It can be observed that our approach did not lose accuracy at all; instead, it gained $0.04\%$, $0.23\%$, and $0.96\%$ of accuracy, while reducing $70\%$ parameters, $50.77\%$ FLOPs, and $53.5\%$ memory in PA, FA, and MA modes respectively. In contrast, only DCP and NPPM reduced $48\%$ and $26\%$ of FLOPs without losing accuracy, while FEAM, RMNv2, and WM could not retain even the baseline accuracy after reducing the complexity. 
\begin{figure*}[ht]
\centering
\includegraphics[width=1\textwidth]{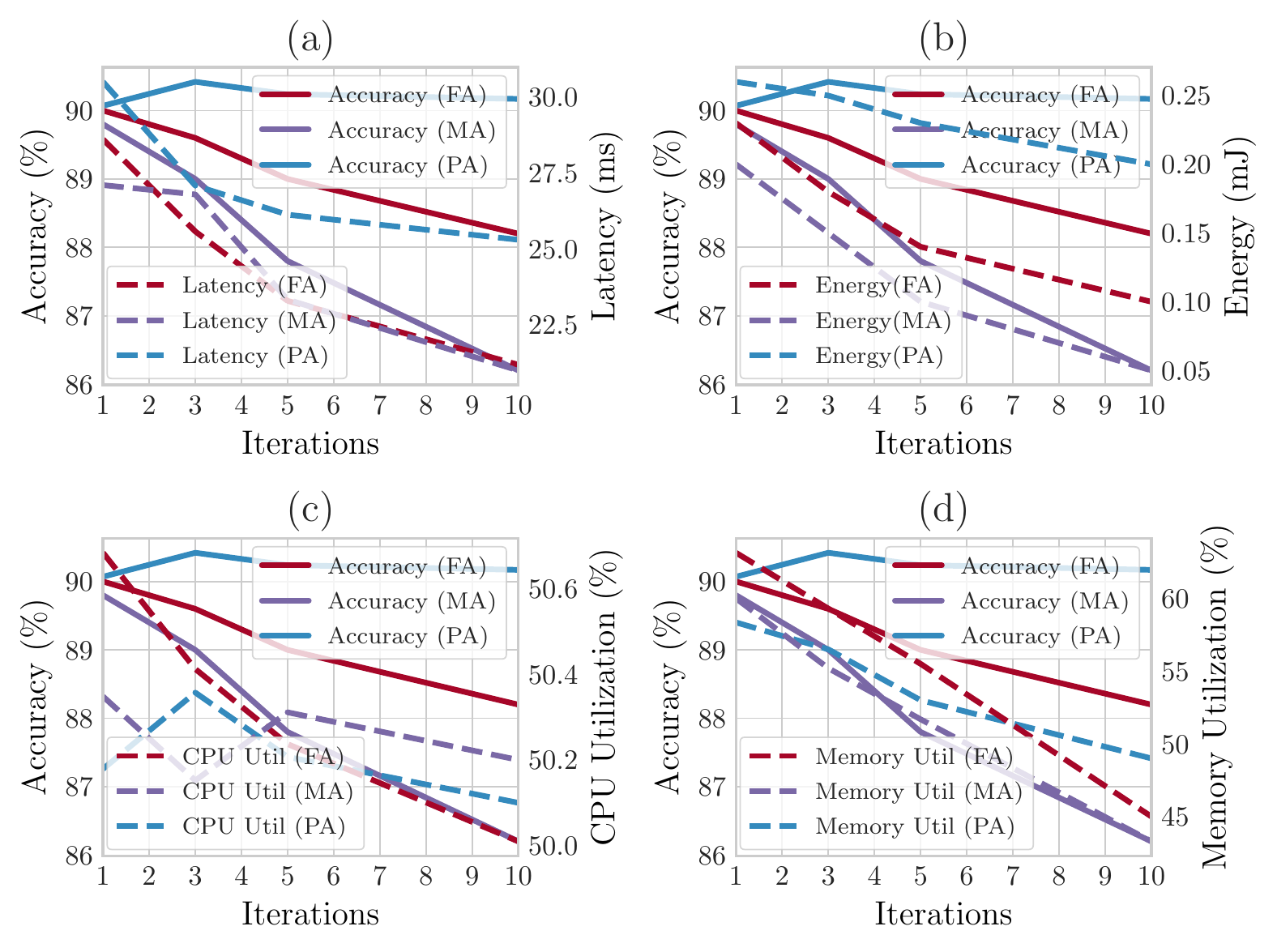}
\caption{Trade-off among accuracy, latency, energy consumption, CPU, and memory utilisation for AlexNet on CIFAR-100.}
\label{fig4}
\end{figure*} 
\subsection{Evaluation on CIFAR-100}
In this subsection, we evaluate the AlexNet and ResNet-50 on CIFAR-100 Dataset. CIFAR-100 also has 60,000 images but is categorized into 100 distinct classes. In this case, the distribution of images per class is 600, of which 500 are used as a training set and the remaining 100 as a testing set. 
\subsubsection{AlexNet}
Similar to VGG-16, AlexNet is also commonly used as a standard CNN for benchmarking model compression approaches. However, the AlexNet architecture is lighter than VGG-16, as it has only 5 convolutional layers, which are followed by 3 fully connected layers. For CIFAR-100, its last layer represents 100 classes having softmax as an activation function, while the remaining layers used the ReLu activation function. For comparison, we report the results of our approach along with IEM\cite{salehinejad2021pruning}, SFP\cite{carballo2020accuracy}, and PLS \cite{jordao2020deep} in Table \ref{tab3}. IEM propose an ising energy model to quantify the inactivity of convolutional kernels in order to prune them. In contrast, SFP proposes a relatively simple approach to assign pruning priority to each layer according to its impact on accuracy. It can be observed in Table \ref{tab3} that our approach achieves a $63.92\%$ reduction in parameters with $0.43\%$ accuracy gain, $79.28\%$ reduction in FLOPs with only $1.15\%$ loss in accuracy, $93.31\%$ reduction in size with only $3.06\%$ loss in accuracy using PA, FA, and MA modes respectively. On the other hand, state-of-the-art single-shot approaches \cite{salehinejad2021pruning, carballo2020accuracy} could not achieve a similar level of reduction and comparable accuracy loss. Moreover, as opposed to our approach, these methods are not versatile as they only focus on single complexity, which is not feasible in production environments. The iterative approach \cite{jordao2020deep} has also failed to gain performance with significant model reduction as our approach. Nevertheless, these approaches are computationally-intensive with complex ranking criteria, which could not contribute to gaining the consistent performance as direct training of pruned model proposed in this work. 

\begin{figure*}[ht]
\centering
\includegraphics[width=1\textwidth]{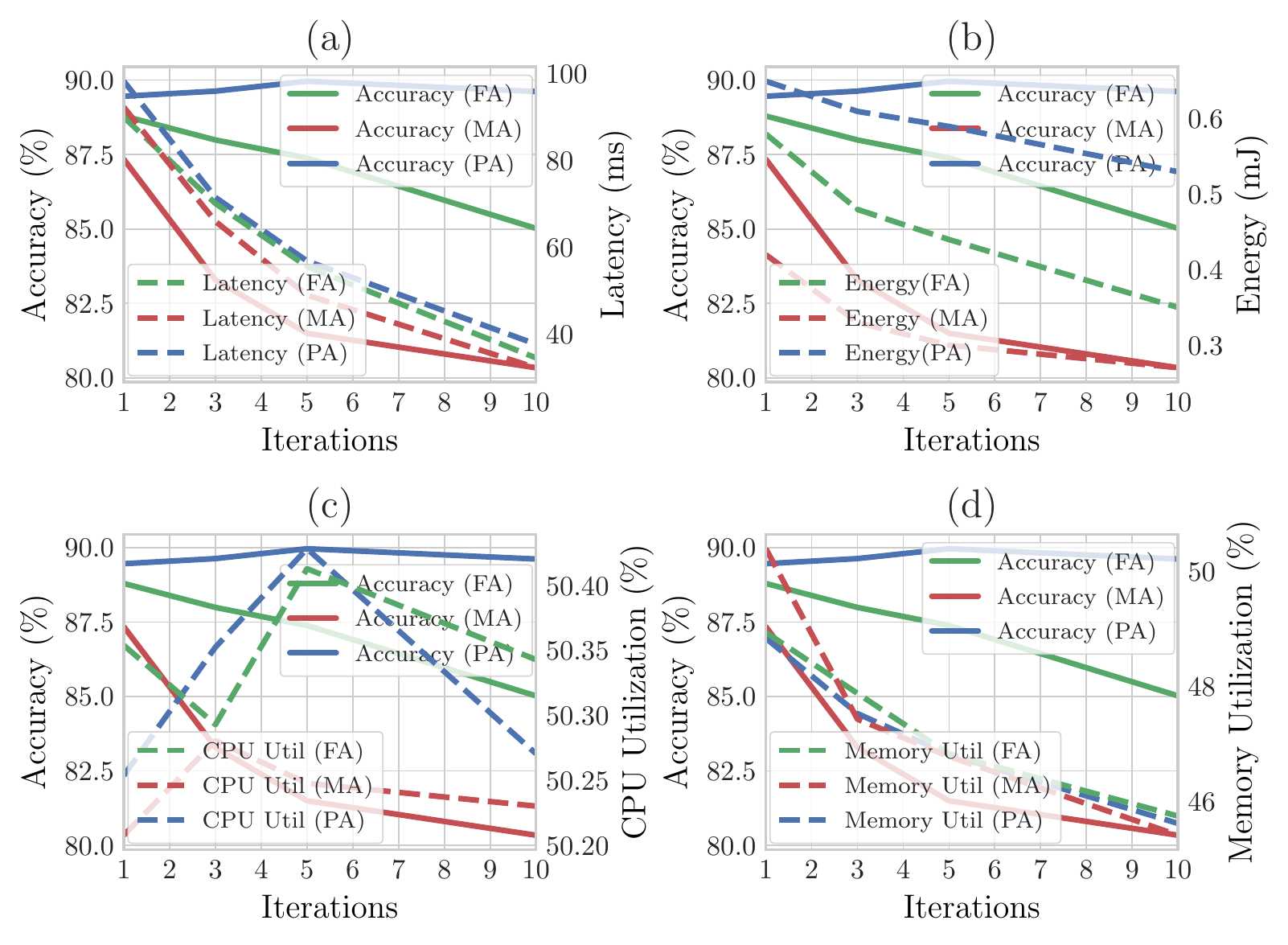}
\caption{Trade-off among accuracy, latency, energy consumption, CPU, and memory utilisation for VGG-16 on CIFAR-10.}
\label{fig5}
\end{figure*} 

\subsubsection{ResNet-50}
ResNet\cite{he2016deep} family is designed with skip connections, similar to MobileNets, to provide high accuracy and faster training. ResNet50 is a 50 layers deep architecture with several residual blocks involving 2 or more $conv$ layers with skip-connections to create a path between 2 residual blocks. A single residual block can be expressed as follows: $y = F(x, \{W_i\}) + x$. In this case, $x$ represents the input vector, and $y$ represents the output vector. $ F(x, \{W_i\})$ corresponds to the residual mapping function and $F + x$ corresponds to the shortcut connection and element-wise addition. Although residual blocks do not strongly impact each other, the dimensions of the feature maps in ResNets must be consistent at the beginning and end of each residual unit. For this reason, pruning residual connections is avoided while designing a pruning strategy. 

Table \ref{tab4} shows the comparative analysis of our approach with DualConv\cite{zhong2022dualconv}, SANet\cite{hacene2021attention}, PCAS\cite{yamamoto2018pcas} and Pruned-B\cite{li2016pruning}. In particular, Pruned-B proposed a one-shot pruning approach by removing those filters which contribute less towards accuracy, while SANet models their pruning approach as finding the right shift for each feature map in $conv$ layers to induce the sparsity in networks, and PCAS develops an attention-based statistical approach to quantify channel importance and prunes the same number of channels from all $conv$ layers. In contrast, DualConv proposes structural modifications by replacing the conventional 3x3 $conv$ operations with stride 1 among all $conv$ layers, apart from the first layer. In Table \ref{tab4}, it can be seen that our approach beats the state-of-the-art mentioned above by reducing $51.33\%$ FLOPs and $30\%$ of memory and gaining $0.34\%$ and $2.23\%$ of accuracy in FA and MA modes respectively. The worst performance was given in PA mode, where $0.25\%$ dropped in accuracy, still competing with the DualConv and Pruned-B.

\begin{table*}
\caption{Evaluation of training time (hours) on Nvidia Tesla K20 GPU for AlexNet and VGG-16 on CIFAR-100 and CIFAR-10 respectively. (\%) shows reduction in \% with respect to baseline.}
\label{tab5}
%%\centering % redundant
%%\renewcommand\footnoterule{\kern -1ex} % What is this instruction doing here??
\renewcommand{\arraystretch}{1.3}
\setlength\tabcolsep{0pt} % make LaTeX figure out the intercolumn separation
\begin{tabular*}{\linewidth}{@{\extracolsep{\fill}} l *{8}{c}} % '1.4', not '6.5'
\toprule 
\rowcolor{Gray2} {Approach} & \multicolumn{3}{c}{AlexNet-CIFAR-100} & \multicolumn{3}{c}{VGG-16-CIFAR-10}\\
\cmidrule{2-4} \cmidrule{5-7} &
Training time & Accuracy (\%) & 
$\Delta$ in Accuracy (\%) & 
Training time & Accuracy{ (\%)} & 
$\Delta$ in Accuracy (\%)  \\
\hline
\textbf{PLS (Baseline)} & $3.46$ & $83.06$ & -- & $3.83$ & $87.05$ & --  \\ 
\textbf{PLS (fine-tuning)} & $1.91$ & $79.65$ & $-4.1$ & $1.91$ & $78.6$ & $-9.7$  \\ 
\textbf{PLS (train+fine-tuning)} & $5.37$ & $79.65$ & $-4.1$ & $5.74$ & $78.6$ & $-9.7$  \\ 
\hline
\textbf{Our (Baseline)} & $2.86$ & $90.28$ & -- & $3.79$ & $90.96$ & -- \\ 
\textbf{Our (PA)} & $2.83 (50.2\%\downarrow)$ & $90.66$ & $0.43$ & $3.37(53\%\downarrow)$ & $92.56 $ & $1.76$  \\ 
\textbf{Our (FA)} & $1.99 (58.96\%\downarrow)$ & $89.24$ & $-1.15$ & $2(65.5\%\downarrow)$ & $87.82$  & $-3.45$   \\ 
\textbf{Our (MA)} & $2.71 (51.4\%\downarrow)$ & $87.51$ & $-3.06$ & $2.41(61\%\downarrow)$ & $82.9 $ & $-8.76$  \\ 
\bottomrule
\end{tabular*}
\end{table*}

\begin{table*}
\caption{Evaluation of training time (minutes) on Nvidia Jetson Nano for AlexNet and VGG-16 on CIFAR-100 and CIFAR-10 respectively. (\%) shows reduction in \% with respect to baseline.}
\label{tab6}
%%\centering % redundant
%%\renewcommand\footnoterule{\kern -1ex} % What is this instruction doing here??
\renewcommand{\arraystretch}{1.3}
\setlength\tabcolsep{0pt} % make LaTeX figure out the intercolumn separation
\begin{tabular*}{\linewidth}{@{\extracolsep{\fill}} l *{8}{c} } % '1.4', not '6.5'
\toprule 
\rowcolor{Gray2}  {Approach} & \multicolumn{4}{c}{AlexNet-CIFAR-100} & \multicolumn{4}{c}{VGG-16-CIFAR-10}\\
\cmidrule{2-5} \cmidrule{6-9} &
Training time & FLOPs & 
Memory (MBs) & Parameters (M) & 
Training time & FLOPs & 
Memory (MBs) & Parameters (M) \\

\hline
\textbf{Baseline} & $7.3$ & $1.72\times10^8$ & $22.34$ & $5.43$ & $9.33$ & $6.29 \times 10^8$ & $62$ & $16.24$ \\ 
\textbf{Our (PA)} & $6.36 (12.8\%\downarrow)$ & $8.8\times10^7 (48\%\downarrow)$ & $7.3(67\%\downarrow)$ & $1.2(78\%\downarrow)$ & $9.28 (0.5\%\downarrow)$ & $4.6 \times 10^8(27\%\downarrow) $ & $41.5(33\%\downarrow)$ & $5(69\%\downarrow)$ \\ 
\textbf{Our (FA)} & $5.26 (27.9\%\downarrow)$ & $2.7\times10^7(84\%\downarrow) $ & $17.4(22\%\downarrow)$& $4.4(18\%\downarrow)$ & $7.46 (20.04\%\downarrow)$  & $2.6\times10^8(58\%\downarrow)$ & $32.6(47\%\downarrow)$ & $8.4(48\%\downarrow)$ \\ 
\textbf{Our (MA)} & $5.85 (19.8\%\downarrow)$ & $1.1\times10^8(35\%\downarrow)$ & $4.2(81\%\downarrow)$ & $2.9(45\%\downarrow)$ & $8.26 (11.4\%\downarrow)$ & $3.7\times10^8 (41\%\downarrow)$ & $34(46\%\downarrow) $ & $8.6(47\%\downarrow)$ \\ 
\bottomrule
\end{tabular*}
\end{table*}

\subsection{Resource-Accuracy Trade-off for Resource Constrained Execution}
It is imperative to evaluate the impact of compression strategies on resource utilization. There is limited attention paid to the end-level benefit of pruning approaches, and the focus has only been on compressing a single complexity with theoretically attractive importance criteria. For example, there is no use in reducing the FLOPs only when executing a microcontroller, as it requires a model with lower memory consumption. Thus, it is critical to answering questions such as \textit{how can we measure the effectiveness of an approach on lesser resource consumption when deployed in production? How can we make a suitable trade-off between different complexities and performance for an optimal deployment?} Several metrics exist to evaluate the suitability of compressed models, such as response time (latency), energy consumption, CPU utilization, and memory utilization. Figs. \ref{fig3} and \ref{fig4} present a comprehensive trade-off between accuracy and different types of resource consumption achieved using our approach in PA, FA, and MA mode for VGG-16 on CIFAR-10 and AlexNet on CIFAR-100, respectively. 

\textbf{Latency.}
Apart from theoretical speedups, i.e., reduction in FLOPs, we also evaluate the practical speedups, i.e., inference time which is also known as latency. We measured the latency as a delay in output required by models to classify an image. Figs. \ref{fig3} and \ref{fig4} show accuracy-latency trade-offs of VGG-16 and AlexNet, respectively. It is clear that we can achieve minimum latency using our approach in MA mode at the cost of significant accuracy loss. However, in FA mode, we can not only minimize latency but also maintain the required accuracy. The inconsistency in FA and MA mode is essentially caused by the unpruned fully connected layers, DRAM access, and the non-parametric layers such as ReLu or Pooling.

\textbf{Energy consumption.}
The energy consumption of a model during inference is considered a critical metric to measure its efficiency in resource-constrained production environments. Based on \cite{han2016eie}, neither the FLOPs nor the number of parameters alone reflects the actual energy consumption of CNNs. Thereby, a model’s total energy consumption is based on energy used in data access and energy required for arithmetic operations on devices. For instance, if each 32-bit FLOP needs 2.3 pJ, then energy for arithmetic operations can be calculated as a product of FLOP counts and 2.3 pJ. Similarly, for data access, retrieving 1MB from DRAM requires 640 pJ, then the product of model size and 640 pJ gives us energy consumption of data access for each model. As shown in Figs. \ref{fig3} and \ref{fig4}, the energy consumption of both  VGG-16 and AlexNet is consistent with the reduction of parameters, FLOPs, and model size. Thus, all three modes are critical in determining the energy usage of a model. Our approach is helpful in such cases, mainly where a developer can compress the model from different aspects, unlike other techniques. Moreover, our approach can help in making the suitable trade-off between energy and accuracy when it comes to low-energy based execution environments such as microcontrollers.

\textbf{CPU and Memory Utilisation.}
Among other key metrics, CPU and memory usage can also be helpful for developers to configure the hardware or cloud infrastructure correctly. When performing a CNN inference, measuring the CPU and RAM usage involves determining the scale of these resources being consumed. The metric for both resources varies from 0\% to 100\%. Practically, the percentage metric is not interesting information itself, but the duration of the resource being used is valuable. Effectively, such information helps developers in managing multiple tasks in parallel if more processing or RAM capacity is available. To meet the tight Quality of Service (QoS) constraints, a developer can compress a CNN in a particular mode based on the usage of CPU and RAM. Figs. \ref{fig3} and \ref{fig4} show the impact of all three modes on both of these resource usages.

\begin{comment}

\begin{table*}\label{tab7}
  \caption{Evaluation on U-Net model with AID and RESIDE datasets . PSNR, SSIM, FLOPs, Memory, and Parameters show respective metrics. $\downarrow$ denotes the reduction in percentage w.r.t the unpruned network.}

\scalebox{1.3}{
  \begin{tabular}{ccccccc}
   \hline
 \textbf{Dataset} &  \textbf{Approach} &  \textbf{PSNR (\%)$\downarrow$} &  \textbf{SSIM (\%)$\downarrow$} &  \textbf{FLOPs (\%)$\downarrow$} &  \textbf{Memory (\%)$\downarrow$} &  \textbf{Parameters (\%)$\downarrow$} \\

    \midrule
    AID & PA & 0.9 & 1.3 & 6.5 & 9.3 & 17.7 \\
& FA & 6.9 & 2.4 & 5.7 & 6.4 & 9 \\
& MA & 8.6 & 5  & 7.5 & 11.9 & 0.9 \\
 \hline
  RESIDE & PA & 1.1 &0.7 &9.1 &7.2 &23.8 \\
& FA & 3.4 &1.2 &11 & 5.2 &6.1 \\
& MA & 4.6 &2.2 & 4.5 & 7.9 &0.9 \\

 \hline
\end{tabular}}
\end{table*}

\begin{figure}
\includegraphics[width=0.5\textwidth]{Result-1.pdf}
\caption{A qualitative trade-off for U-Net on AID using PA, FA, and MA modes.}
\label{fig5}
\end{figure} 

\begin{figure}
\includegraphics[width=0.5\textwidth]{Result-2.pdf}
\caption{A qualitative trade-off for U-Net on RESIDE using PA, FA, and MA modes.}
\label{fig6}
\end{figure} 
\end{comment}
\subsection{Training Efficiency}
In this subsection, we evaluate the proposed approach in terms of training efficiency on a resource-rich GPU and a resource-constrained GPU to justify the key idea of reducing the training bottleneck and achieving competitive performance. 
\subsubsection{Evaluation on Resource-rich GPU}

In Table \ref{tab5}, we show the comparison of our approach with a conventional iterative pruning approach, i.e., PLS \cite{jordao2020deep}. Most of the conventional approaches are alike since the methodology follows the training-pruning-fine-tuning procedure. The resource-rich GPU used for training is Nividia Tesla K20. We follow the training parameters originally mentioned in \cite{jordao2020deep} for training the baseline model and fine-tuning the pruned model after 10 iterations. The models are trained for 150 epochs before and after pruning. Hence, in the case of PLS, the baseline (unpruned) AlexNet took $3.46$ hours, while the retraining/fine-tuning of the pruned model took $1.91$ hours. Therefore, the total time in achieving the pruning objective required $5.37$ hours which indicates the overall training cost by compromising the $4.1\%$ of accuracy. In contrast, our approach reduced the training time by $50.2\%$ in PA mode, $59.96\%$ in FA mode, and $51.4\%$ in MA mode, along with maintaining competitive accuracy. This kind of training efficiency can be achieved with our approach as it relies on training the pruned model directly instead of following the training-pruning-fine-tuning procedure.

\subsubsection{Evaluation on Resource-constrained Edge device}
In Table \ref{tab6}, we show the training performance of our approach on a relatively resource-constrained edge device, i.e., Nvidia Jetson Nano, which is often utilized for edge AI applications. This edge device comprises a GPU with 128 NVIDIA CUDA® cores, a Quadcore ARM Cortex-A57 CPU, 4GB of RAM, and 16GB of storage. Traditionally, due to its resource-constrained nature, this device is used to facilitate deep learning inference using pre-deployed models trained on a certain application. However, in this article, we consider this device only for benchmarking the training efficiency of our approach. Since the device has very sparse resources, training 150 epochs for each model would require a lot of time. Thus, for the sake of the evaluation, we trained the AlexNet and VGG-16 for only 1 epoch, with a batchsize of 5, on $20\%$ training data of CIFAR-100 and CIFAR-10, respectively. In each pruning mode, we prune the models upto 50 iterations and report the obtained results in Table \ref{tab6}. It can be seen that, for both AlexNet and VGG-16, FA mode has performed best in terms of reducing the training time. Since the objective of FA mode is to identify and prune those layers which consume most of the FLOPs, it eventually reduces the number of operations performed in the forward and backpropagation. Hence, our approach can benefit resource-constrained edge devices to perform a fractional amount of training as well, such as for the purpose of fine-tuning fresh data or contributing to a federated learning task. 

\section{Conclusion}
In this paper, we present a complexity-driven structured pruning, which enables compression of CNNs in different modes, i.e., parameter-aware, FLOPs-aware, and memory-aware. It is obvious from the above discussion that every layer of a CNN not only shows a different nature of complexity but also contributes differently to overall model complexity. Thus, unlike state-of-the-art approaches, our proposed work takes care of both these aspects when pruning a certain model. The proposed pruning scheme is computationally efficient since every model is first pruned and then directly trained instead of typical three-stage pipelines. As shown in the results, the proposed method can accelerate CNNs in different pruning modes without losing much accuracy. Moreover, developers can benefit from this approach by compressing CNNs in different modes to trade accuracy with various resources. In the future, we aim to utilize this approach for distributed execution on heterogeneous edge devices for real-time inference.

\section*{Acknowledgement}
This  research  was  supported  by  Science  Foundation  Ireland  and  the  Department  of  Agriculture,  Food  and Marine on behalf of the Government of Ireland VistaMilk research centre under the grant 16/RC/3835.

\ifCLASSOPTIONcaptionsoff
  \newpage
\fi
\bibliographystyle{IEEEtran}
\bibliography{main}

\end{document}